# SPARK: Scalable Real-Time Point Cloud Aggregation with Multi-View Self-Calibration

Chentian Sun

**Abstract—Real-time multi-camera 3D reconstruction is a key problem in 3D perception, immersive interaction, and robotic systems. Although modern depth cameras and multi-view reconstruction techniques can capture high-resolution depth data at high frame rates, reliably and efficiently fusing multi-view depth information into high-quality point clouds under real-time constraints remains challenging. In particular, multi-view point cloud fusion, camera extrinsic uncertainty, and the computational scalability of multi-camera point clouds constitute major bottlenecks for achieving truly large-scale real-time 3D reconstruction.**

**In this paper, we propose SPARK, a self-calibrating real-time multi-camera point cloud reconstruction framework that jointly models multi-view point cloud fusion and camera extrinsic uncertainty within a unified system. Unlike conventional methods that rely on accurate pre-calibration, global volumetric representations, or iterative optimization, SPARK performs frame-wise fusion directly in the point cloud domain and achieves linear scalability with respect to the number of cameras through a modular design. Specifically, the framework consists of two key components:**

**(1) A geometry-aware online extrinsic estimation module that leverages implicit geometric representations learned by multi-view reconstruction networks as priors, and achieves stable calibration-free extrinsic self-correction through cross-view reprojection consistency and temporal constraints;**

**(2) A confidence-driven point cloud generation and fusion strategy that jointly models depth reliability and visibility at both pixel and point levels, effectively suppressing noise and view-dependent inconsistencies.**

**Benefiting from frame-wise independent point cloud representations and an accumulation-free fusion scheme, SPARK can continuously produce stable point cloud streams in dynamic scenes, while its computational and memory costs grow linearly with the number of cameras, thereby enabling large-scale real-time reconstruction with hundreds of millions of points. Extensive experiments on real-world multi-camera systems demonstrate that the proposed method outperforms representative existing approaches in terms of extrinsic accuracy, geometric consistency, temporal stability, and real-time performance, validating its effectiveness and scalability in practical multi-camera settings.**

## I. INTRODUCTION

Real-time multi-view 3D reconstruction is a fundamental problem in computer vision, playing a crucial role in applications such as virtual and augmented reality, robotic navigation, and immersive interaction. With the rapid development of depth cameras and multi-view stereo techniques, modern systems can capture high-resolution depth maps from multiple viewpoints at high frame rates. However, under real-time constraints, reliably and efficiently fusing multi-view depth information into high-quality point clouds remains challenging. In particular, multi-view point cloud fusion, extrinsic parameter uncertainty, and the computational scalability of multi-camera point clouds constitute major bottlenecks for achieving truly large-scale real-time 3D reconstruction.

Existing 3D reconstruction methods are typically limited by their design principles, addressing only one or two of these challenges, and are rarely able to handle multi-view fusion stability, extrinsic uncertainty, and computational scalability within a unified framework. Broadly, multi-view reconstruction methods can be categorized into three types. The first type relies on voxel or implicit volumetric fusion (e.g., TSDF or occupancy grids), which accumulates depth observations in a global voxel space. While this approach can achieve high precision under controlled conditions, it depends on accurate and static extrinsic calibration, and its computational and memory costs scale cubically with resolution and scene size, making it difficult to extend to high-resolution, multi-camera real-time scenarios. The second type consists of SLAM-based real-time point cloud reconstruction methods, which generate continuous point clouds through frame-to-frame registration and pose optimization. In multi-camera settings, these methods require additional cross-view constraints or global optimization, leading to increased computational complexity and vulnerability to error accumulation, causing extrinsic drift and geometric distortions. The third type includes multi-frame joint optimization and offline reconstruction methods, such as NeRF or 3D Gaussian Splatting. Although capable of producing high-quality reconstructions, these approaches typically assume known extrinsics or rely on long-term iterative optimization, making them computationally intensive and unsuitable for stable real-time output under multi-camera, high-frame-rate conditions.

In real multi-camera systems, depth observations are further affected by sensor noise, viewpoint changes, occlusions, and reflective or transparent surfaces, which often lead to cross-view inconsistencies and local artifacts. Meanwhile, high-resolution multi-camera inputs significantly increase the computational and storage burden, limiting the applicability of methods that rely on global optimization or high-dimensional representations. Moreover, existing systems typically depend on precise offline extrinsic calibration; even minor calibration errors or camera displacements can introduce noticeable ghosting and geometric distortions, severely affecting stability. Therefore, multi-view fusion, extrinsic uncertainty, and scalable computation remain key bottlenecks for achieving practical, large-scale real-time 3D reconstruction.

To address these challenges, we propose SPARK, a scalable self-calibrating framework for real-time multi-camera point cloud reconstruction. Unlike voxel based, SLAM-accumulation, or offline optimization methods, SPARK performs multi-view fusion directly in the point cloud domain and represents data as frame-wise, independent point cloud fragments for efficient computation. The framework further introduces a geometry-aware online extrinsic estimation module, which leverages the implicitly learned geometric structures from multi-view reconstruction networks as a prior, and enforces cross-view reprojection and temporal consistency constraints to achieve stable self-calibration without manual intervention. In addition, a confidence-driven point cloud generation and fusion strategy explicitly models measurement reliability during depth back-projection and fusion, effectively suppressing noise and cross-view inconsistencies.

Since both point cloud generation and fusion are performed independently at the frame level, SPARK exhibits linear computational complexity with respect to the number of cameras, naturally supporting multi-camera scalability. It can continuously output stable point cloud streams in dynamic scenes without accumulating errors. Extensive experiments on real multi-camera datasets demonstrate that SPARK consistently outperforms existing methods in reconstruction accuracy, temporal stability, and real-time performance.

The main contributions of this work are summarized as follows:

(1) A geometry-aware online extrinsic estimation module that leverages implicit geometric representations learned by multi-view reconstruction networks as priors, and achieves stable calibration-free extrinsic self-correction through cross-view reprojection consistency and temporal constraints;

(2) A confidence-driven point cloud generation and fusion strategy that jointly models depth reliability and visibility at both pixel and point levels, effectively suppressing noise and view-dependent inconsistencies.

Extensive experiments on real multi-camera datasets validate the effectiveness of the proposed framework in terms of accuracy, stability, and scalability.

## II. Related Work

Real-time multi-camera 3D reconstruction systems involve two core problems:

(1) Extrinsic estimation for multi-view alignment;

(2) Multi-view 3D reconstruction and fusion.

These topics have long been a focus in computer vision, and extensive research has been conducted in both areas.

### 2.1 Extrinsic Estimation and Online Self-Calibration

Accurate and stable extrinsic estimation is essential for multi-camera fusion. However, under high frame rates, dynamic scenes, and multi-stream inputs, traditional methods often struggle to achieve both real-time performance and robustness. Existing research mainly falls into the following directions:

#### 2.1.1 Traditional Calibration and Optimization

Offline calibration (e.g., multi-view checkerboard) provides high-precision static extrinsics but assumes fixed camera poses, which is unsuitable for dynamic scenarios or long-term deployment, potentially causing drift. Online geometric optimization attempts to adjust extrinsics during runtime, as in SLAM/VIO frameworks. For example, ORB SLAM3 [Campos et al., 2021] jointly optimizes system states and extrinsics. However, these approaches often rely on global bundle adjustment (BA), which is computationally expensive and difficult to scale to large multi-camera systems.

#### 2.1.2 Learning-Based Extrinsic Prediction

In recent years, deep neural networks have been increasingly applied to directly predict camera poses and extrinsics.

In 2024, VGGSfM leverages multi-view geometric features through an end-to-end network to infer camera poses. DFSFM integrates deep feature representations with iterative optimization for robust multi-view pose estimation. PoseDiffusion employs diffusion models to progressively refine camera poses, providing high-precision predictions and uncertainty estimates.

These methods demonstrate the potential of learning-based approaches for visual odometry and extrinsic estimation, reducing reliance on precise calibration and global bundle adjustment. However, most remain designed for small-scale or controlled settings and are not easily scalable to large-scale, multi-camera, real-time systems.

#### 2.1.3 Multi-View Reconstruction with Large Models

With the development of Vision Large Models (VLMs) and Transformer architectures, recent work has explored using large pre-trained models for multi-view scene understanding and 3D reconstruction. Methods such as VGGT, MapAnything, and DepthAnything V3 leverage cross-view attention to aggregate multi-view features, enabling joint depth prediction and preliminary extrinsic estimation without relying on precise calibration. These approaches are robust to sparse views and occlusions, but a unified system for online multi-camera extrinsic self-calibration is still lacking.

In light of these limitations, we propose the GMAC module, which leverages multi-view geometric priors to perform online self-calibration of multi-camera extrinsics. It integrates naturally with real-time point cloud fusion, improving both reconstruction accuracy and system scalability. Moreover, GMAC is complementary to existing multi-view reconstruction models, enhancing both real-time extrinsic calibration accuracy and overall multi-view reconstruction performance.

### 2.2 Multi-View 3D Reconstruction and Fusion

Multi-view 3D reconstruction aims to generate high-quality 3D representations from multi-source depth or visual inputs. Existing methods can be broadly categorized into voxel based approaches, SLAM-Driven point cloud reconstruction, and learning-based neural approaches.

#### 2.2.1 Voxel / Occupancy-Based Methods

Classic voxel fusion methods integrate depth frames using TSDF or occupancy grids to achieve real-time scene reconstruction, such as certain Tesla occupancy-based methods and other TSDF voxel fusion approaches. While these methods perform well in static scenes, they suffer from several limitations:

(1) The computational cost of global voxel grids increases rapidly with the number of cameras and scene complexity;

(2) They are sensitive to extrinsic errors, with drift potentially causing misalignment;

Their support for dynamic scenes is limited.

#### 2.2.2 SLAM-Driven Point Cloud Reconstruction

Real-time SLAM systems generate continuous point clouds through frame-to-frame registration and global optimization. Representative examples include KinectFusion [Newcombe et al., 2011], ElasticFusion [Whelan et al., 2015], and NICE SLAM [Zhu et al., 2022]. These methods combine voxel or optimization structures to maintain spatial consistency, but they typically rely on global bundle adjustment (BA) to preserve geometric accuracy, leading to high computational cost and limited scalability to multi-camera setups.

#### 2.2.3 Neural / Implicit Representation Methods

Learning-based neural approaches generate high-quality 3D reconstructions through implicit representations, with representative examples including NeRF [Mildenhall et al., 2020] and 3DGS. These methods achieve superior reconstruction accuracy but generally assume static scenes and precise extrinsics, and they require substantial optimization time, making them unsuitable for real-time multi-camera output.

Overall, existing approaches still involve trade-offs among real-time performance, scalability, and robustness in dynamic scenes. The point cloud generation and fusion strategy proposed in this work leverages pixel-level and point-level confidence modeling, visibility-driven weight assignment, and frame-level independent, non-accumulated fusion. This approach effectively suppresses noise and multi-view depth conflicts, enabling stable and efficient real-time multi-camera reconstruction. By integrating automatic multi-camera extrinsic estimation, the proposed SPARK framework naturally supports large-scale multi-camera systems.

## III. Proposed Methods

### 1. Problem Definition and Overall Framework

Consider a multi-camera system consisting of $N$ cameras. At time step $t$, the system receives synchronized RGB or depth data:

$$\left\{I_i^t, D_{i\ i}^t\right\}_{i=1}^N \tag{1}$$

where $D_i^t$ denotes the raw depth observation of the $i$-th camera, and the intrinsic parameters $K_i$ are known. The extrinsic parameters, however, are unknown or may contain errors:

$$T_i^t = \begin{bmatrix} R_i^t & p_i^t \\ 0 & 1 \end{bmatrix} \in SE(3) \tag{2}$$

where $R_i^t \in \mathbb{R}^{3\times3}$ is the rotation matrix, $p_i^t \in \mathbb{R}^{3\times3}$ is the translation vector, the superscript $t$ indicates the time step.

The system aims to estimate stable extrinsics $\{T_i^t\}$ in real time without manual calibration, and generate high-quality multi-view fused point clouds $p_i^t$ with linear time complexity**.**

The proposed SPARK framework consists of three key components:

(1) Online extrinsic estimation based on implicit multi-view geometric representations;

(2) Confidence-aware frame-level point cloud generation;

(3) Dynamic multi-view point cloud fusion.

This design avoids global voxel grids, long-term accumulation, or iterative optimization, naturally supporting dynamic scenes and scalable multi-camera setups.

### 2. Online Extrinsic Estimation Based on Implicit Multi-View Geometry

Accurate extrinsic calibration is critical for high-quality multi-view point cloud fusion in multi-camera systems. Traditional methods rely on manual calibration or single-frame optimization, which struggle to meet the requirements of real-time dynamic scenes. To address this, we propose an online extrinsic estimation method based on implicit multi-view geometry (GMAC method), which predicts initial extrinsics from shared latent features and refines them under geometric and temporal constraints to achieve stable, real-time estimation.

GMAC consists of three components: implicit geometric prior modeling, lightweight multi-view feature extraction, and geometric and temporal consistency constraints. The following sections describe each of these components in detail.

#### 2.1 Implicit Geometric Prior Modeling

Multi-view reconstruction networks implicitly learn cross-view geometric consistency during training. Let the multi-view feature extraction be:

$$F^t = \phi(\{I_i^t\}) \tag{3}$$

where $F^t$represents the latent geometry features shared across views. Based on these features, the camera extrinsics are modeled as a function of the latent representation:

$$\{\mathrm{T}_{\mathrm{i},0}^{\mathrm{t}}\}_{\mathrm{i}=1}^{\mathrm{N}} = \mathrm{g}(\mathrm{F}^{\mathrm{t}}) \tag{4}$$

where $g(\cdot)$ is a lightweight extrinsic regression head that outputs the initial extrinsic estimate. Unlike per-camera regression, GMAC introduces multi-view geometric coupling through shared features, enabling extrinsic prediction to rely on global consistency.

#### 2.2 Network Architecture Optimization and Lightweight Design

Directly using the full reconstruction network for extrinsic estimation may introduce irrelevant computations (e.g., explicit 3D representation or rendering branches), reducing prediction stability. To address this, the network structure is optimized:

(1) Retain multi-view feature extraction and interaction modules;

(2) Remove explicit 3D reconstruction and rendering branches;

(3) Add a lightweight extrinsic regression head on shared features.

This design ensures that extrinsic estimation focuses on modeling multi-view geometric relationships, improving robustness while reducing real-time computational overhead.

2.3 Geometric and Temporal Consistency Constraints

The initial predictions $\left[T_{i,0}^{t}\right]$ are further refined using geometric and temporal constraints:

(1) Reprojection consistency constraint:

$$L_{geo} = \sum_{i,j} \left\| \pi(T_i^t, X) - \pi(T_j^t, X) \right\|_2^2 \tag{5}$$

where $\pi(\cdot)$ denotes the camera projection function, and $X$ represents observable or implicit 3D points. This constraint encourages different cameras to produce consistent projections for the same scene point.

(2) Reprojection consistency constraint:

$$L_{net} = \sum_i \left\| \widehat{T_i^t} - T_i^t \right\|_2^2 \tag{6}$$

The network predictions serve as a data-driven geometric prior, guiding optimization within a reasonable local neighborhood.

(3) Temporal consistency constraint:

$$L_{temp} = \sum_t \| T_i^t - T_i^{t-1} \|_2^2 \tag{7}$$

This ensures smooth extrinsic estimates across consecutive frames and suppresses noise-induced jitter.

The joint optimization objective is:

$$argmin_{\{T_i^t\}} (L_{net} + \lambda L_{geo} + \mu L_{temp}) \tag{8}$$

where $\lambda$, $\mu$ are weighting hyperparameters.

After joint optimization under geometric and temporal consistency constraints, the stable online extrinsic is obtained, denoted as $\hat{T}_i^t$, and is used for subsequent multi-view point cloud generation and fusion.

3. Large-Scale Point Cloud Fusion and Generation Based on Confidence

Based on the real-time and accurate multi-camera extrinsics $\hat{T}_i^t$ obtained in the previous section using the GMAC method, SPARK evaluates the reliability of each depth measurement through pixel-level confidence estimation and performs weighted fusion of multi-view observations according to geometric visibility. This enables frame-wise, stateless, and linear-complexity large-scale real-time point cloud generation and fusion. Unless otherwise specified, all variables in this section are defined at the current time step $t$, and the time index is omitted for notational simplicity.

The following sections describe pixel-level confidence modeling, frame-level point cloud generation based on confidence, and the process of dynamic viewpoint weight allocation and multi-view point cloud fusion.

3.1 Measurement Confidence Modeling

For each pixel $(x, y)$, a scalar confidence value is assigned:

$$C_i(x, y) \in [0,1] \tag{9}$$

which reflects the geometric reliability of the corresponding depth observation in 3D reconstruction. Instead of modifying or refining depth values, FUSE-Flow explicitly preserves measurement uncertainty and suppresses unreliable observations in later stages through confidence-aware processing.

The measurement confidence is computed by jointly considering the following factors:

(1) Depth gradient $G(x, y)$ : regions with large depth variations are more likely to be affected by occlusions or matching errors;

(2) Local depth consistency $\sigma_{local}(x, y)$ : smaller depth variance in a local neighborhood indicates higher measurement stability;

The confidence is defined as:

$$C_i(x, y) = \alpha \cdot \frac{1}{1+\beta G(x,y)} + \gamma \cdot \frac{1}{1+\delta \sigma_{local}(x,y)} \tag{10}$$

Where α, β, γ, δ are weighting parameters. In this work, $\alpha$ and $\beta$ are set to 0.5, and $\gamma$ and $\delta$ are set to 1.

3.2 Visibility Confidence Modeling

In FUSE-Flow, $V_i(x, y)$ denotes the Visibility confidence of the 3D point corresponding to pixel $(x, y)$ for camera $i$. It is used to determine whether the 3D point lies within the valid field of view of the camera and is not occluded. Its computation is generally performed in two steps:

(1) Field-of-View Check

The 3D point $P(x, y)$ is projected onto the image plane of camera $i$:

$$u_i = K_i T_i^{-1} P(x, y) \tag{11}$$

The projected coordinates $u_i$ are then checked to see if they fall within the image boundaries. For points $\acute{u}_i$ that fall within the image boundaries, the 3D point is potentially visible.

(2) Occlusion Check

For 3D points that pass the field-of-view check, occlusion is determined through depth comparison. The visibility mask is computed as:

$$V_i(x, y) = \begin{cases} 1, if\ Z_i(P(x, y)) < D_i(\acute{u}_i) \\ 0, otherwise \end{cases} \tag{12}$$

where $Z_i(P(x,y))$ is the depth of the 3D point $P(x,y)$ in the coordinate system of camera $i$, and $D_i(\acute{u}_i)$ is the depth value at the projected pixel $\acute{u}_i$ in the depth map of camera $i$.

3.3 Frame-level Point Cloud Generation

In FUSE-Flow, each depth pixel is directly back-projected into 3D space using the camera geometry, without any explicit depth refinement:

$$P_i(x,y) = T_i \cdot K_i^{-1} \cdot (D_i^r(x,y) \cdot u_h), u_h = [x,y,1]^T \quad (13)$$

where $P_i(x,y) \in \mathbb{R}^3$ represents the 3D point corresponding to pixel $(x,y)$, $u_h$ is the homogeneous coordinate vector of the pixel $(x,y)$, $D_i^r$ is the depth value obtained after necessary preprocessing (such as filtering, screening, etc.) of the original depth value.

All back-projection operations are performed in parallel on the GPU at the pixel level, resulting in a point cloud generation process whose computational complexity is strictly linear with respect to the number of pixels. This allows the FUSE-Flow system to generate tens of millions of 3D points per frame in real time.

To further improve robustness and efficiency, a confidence-based gating mechanism is applied during point cloud generation:

$$\tilde{P}_i(x,y) = \begin{cases} P_i(x,y), & C_i(x,y) > \tau \\ 0, & otherwise \end{cases} \quad (14)$$

Here, $\tau$ is the threshold of Measurement confidence. Low-confidence 3D points are suppressed during the generation stage, thereby reducing noise and lowering the computational burden. In this work, $\tau$ is set to 0.6 to suppress noise while retaining valid 3D points.

3.4 Dynamic Viewpoint Weight Allocation

In multi-camera systems, the same spatial region is often observed from multiple viewpoints with varying geometric reliability due to occlusions, viewing angles, and sensor noise. Treating all observations equally may introduce fusion artifacts, especially near occlusion boundaries.

To address this issue, FUSE-Flow introduces a dynamic viewpoint weight allocation mechanism. First, the point cloud fragments from each camera, $\tilde{P}_i(x,y)$ are transformed into a unified 3D coordinate system. For overlapping spatial regions, the observations of each 3D point from different cameras are collected, and a weight is assigned to each camera's observation:

$$\omega_i(x,y) = \frac{C_i(x,y) \cdot V_i(x,y)}{\sum_{j=1}^{N} C_j(x,y) \cdot V_j(x,y)} \quad (15)$$

where $C_i(\mathrm{x},\mathrm{y})$ is the Measurement confidence of the spatial point $X_k$ derived from the corresponding pixel in $\tilde{P}_i(x,y)$, and $V_i(X_k)$ is a Visibility confidence indicating whether $X_k$ is within the valid field of view of camera $i$. Visibility is determined via ray casting to check for geometric occlusions.

This mechanism ensures that only reliable and visible observations contribute significantly to the fusion, effectively reducing errors caused by occlusions and viewpoint conflicts.

3.5 Multi-view Point Cloud Fusion

For a 3D point at the same spatial location, its valid observations from different cameras form the set $\{\tilde{P}_i(x,y)\}$. Fusion is performed using a confidence-weighted average:

$$P(x,y) = \sum_{i=1}^{N} \omega_i(x,y) \tilde{P}_i(x,y) \quad (16)$$

By unifying per-point confidence modeling with geometric visibility constraints, FUSE-FLOW achieves linear-complexity, real-time multi-view point cloud fusion without voxel representations, providing an efficient and robust solution for large-scale multi-camera 3D reconstruction.

## IV. Experimental Results

4.1 Experimental Setup

To validate the effectiveness and practicality of the proposed SPARK self-calibrating multi-camera real-time 3D reconstruction framework in complex real-world scenarios, a comprehensive system-level experimental evaluation is conducted. The experiments focus on three core aspects:

(1) Accuracy and stability of online multi-camera extrinsic estimation;

(2) Geometric consistency of multi-view point cloud reconstruction;

(3) Real-time performance and system scalability.

The experiments are conducted on a multi-camera RGB-D system consisting of 1–8 fixed, synchronized cameras covering multiple viewpoints. The test scenarios include:

(1) Structured indoor scenes: featuring regular geometric structures such as walls, tables and chairs, and corridors;

(2) Complex geometric scenes: with large viewpoint overlaps and rich geometric details;

(3) Dynamic scenes: containing non-rigid objects such as pedestrians, hand movements, and movable items.

Each type of scene ensures multi-view coverage and varied motion states, allowing a thorough evaluation of the system's performance under both static and dynamic conditions.

Considering that the SPARK system is composed of two main modules—extrinsic estimation and point cloud reconstruction—we evaluate the performance of each module separately.

4.2 Online Multi-Camera Extrinsic Estimation Results

4.2.1 Comparison Methods and Evaluation Metrics on Extrinsic Estimation Module

In the extrinsic estimation experiments, VGGSfM [6], MapAnything [7], and VGGT [8] were selected as the backbone networks for multi-view reconstruction. In these experiments, only structural pruning and functional reconfiguration of the networks were performed, without any retraining, in order to verify the backbone-agnostic nature and transferability of the proposed method. This section evaluates the performance of SPARK's extrinsic estimation module, GMAC, in multi-camera extrinsic calibration, including overall accuracy and stability, as well as the contributions of its individual components.

TABLE I
THE PERFORMANCE OF GMAC WITH DIFFERENT BACKBONE

| Backbone | Method | Structured | Complex | Dynamic |
|---|---|---|---|---|
| VGGT | Original | 2.12 / 0.061 / 0.014 | 2.28 / 0.065 / 0.015 | 2.91 / 0.078 / 0.021 |
| | GMAC | 1.22 / 0.034 / 0.005 | 1.30 / 0.038 / 0.006 | 1.51 / 0.044 / 0.008 |

TABLE II
THE PERFORMANCE OF EACH COMPONENT IN GMAC

| Scenes / Ablation Module | Structured | Complex | Dynamic |
|---|---|---|---|
| No Geometric Consistency | 1.78 / 0.048 / 0.011 | 1.94 / 0.051 / 0.013 | 2.03 / 0.055 / 0.015 |
| SPARK | 1.22 / 0.034 / 0.005 | 1.30 / 0.038 / 0.006 | 1.51 / 0.044 / 0.008 |

The evaluation metrics consist of two parts:

(1) Accuracy of online multi-camera extrinsic estimation, including both translation and rotation accuracy;

(2) Stability of multi-camera extrinsic estimation, computed based on a frame-difference method.

4.2.2 Extrinsic Estimation Accuracy and Stability

To evaluate the effectiveness of the GMAC method, we tested different backbone networks in various scenarios, and the results are shown in Table I.

As shown in Table I, across different real-world environments, the proposed GMAC method consistently improves the accuracy of extrinsic parameter prediction when applied to various backbone networks, providing a reliable foundation for subsequent point cloud generation and fusion. The performance comparison with VGGSfM and MapAnything will be released in version v4.

4.2.3 Ablation Study Analysis

To verify the individual contributions of GMAC's modules to extrinsic parameter estimation, we conducted ablation experiments on the VGGT backbone network by removing the geometric consistency constraint, network prior constraint, and temporal consistency constraint respectively. The results are shown in Table II.

From Table II, several observations can be made:

(1) Geometric consistency constraint contributes the most to extrinsic parameter accuracy, as removing it leads to a significant increase in both rotation and translation errors.

(2) Network prior constraint helps restrict the optimization space and prevent unstable solutions; its removal slightly increases the errors.

(3) Temporal consistency constraint primarily suppresses inter-frame jitter, with a particularly noticeable effect in dynamic scenes.

We also evaluated the impact of progressively removing GMAC components on the qualitative reconstruction results, as illustrated in Fig. 1. It can be observed that the original VGGT method predicts generally accurate extrinsic parameters, but noticeable deviations remain in fine details. After incorporating GMAC, the overall accuracy of extrinsic estimation is significantly improved. When the geometric consistency constraint is removed, the reconstruction quality degrades the most compared with the full GMAC model. Removing the network prior constraint also leads to a noticeable performance drop. Overall, the complete method achieves the best results across all three scene types, further validating the effectiveness of each component in GMAC.

The ablation experiments on the components related to Network Prior and Temporal Consistency will be conducted for the release of the v4 version.

4.3 Multi-View Point Cloud Reconstruction and Fusion Results

4.3.1 Comparison Methods and Evaluation Metrics on Point Cloud Reconstruction

Considering the significant differences in design assumptions among existing 3D reconstruction methods—such as the number of cameras, scene dynamics, real-time capability, and the requirement for offline extrinsic calibration—directly comparing them under a unified setup is neither fair nor technically feasible.

In particular, the combination of multi-camera, dynamic scenes, real-time reconstruction, and online extrinsic estimation is currently unsupported by most existing methods.

To fairly and effectively evaluate the proposed SPARK method, we adopt a capability-aligned comparison principle, performing experiments only within the operational scope of each method rather than enforcing a unified setting.

TABLE III
COMPARISON OF SUPPORTED FEATURES AMONG DIFFERENT 3D RECONSTRUCTION METHODS

| Features / Methods | Multi Camera | Dynamic Scenes | Real Time | Online Extrinsics |
|---|---|---|---|---|
| 3DGS | √ | Δ | × | × |
| ElasticFusion | × | × | √ | √ |
| PatchmatchNet | √ | × | × | × |
| DynamicFusion | × | √ | √ | × |
| R3D3 | √ | √ | Δ | × |
| SPARK | √ | √ | √ | √ |

√: Supported ×: Not supported Δ: Marginally supported

We categorize the experiments into four typical scenarios:

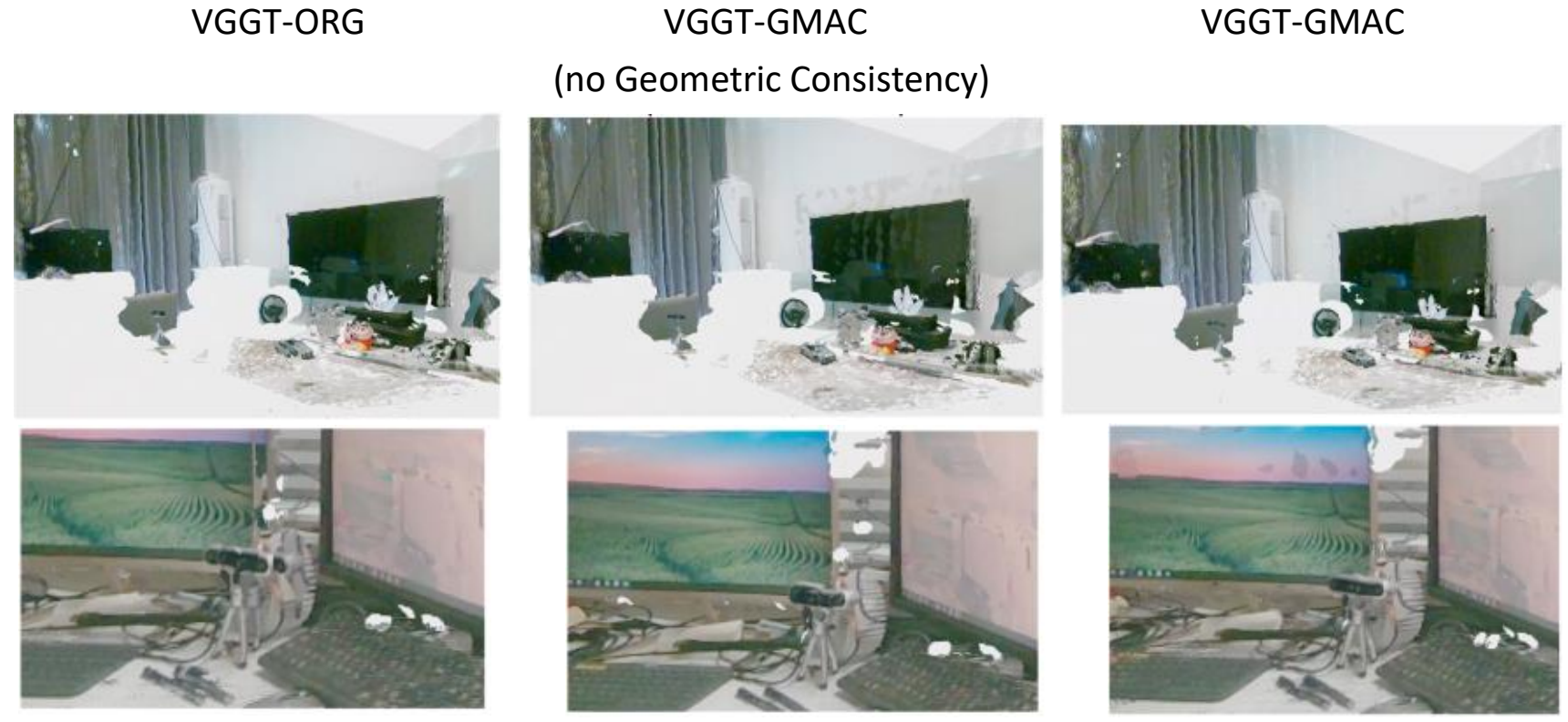

Fig. 1. The original extrinsic parameter estimation performance of VGGT prediction versus the performance after sequentially removing relevant components of GMAC

(1) Single-camera static scenes: 3DGS [18], ElasticFusion [20]
(2) Single-camera dynamic scenes: DynamicFusion [22]
(3) Multi-camera static scenes: PatchmatchNet [21]
(4) Multi-camera dynamic scenes: R3D3 [23]

The capability boundaries of 3D reconstruction methods are summarized in Table III.

It can be observed that SPARK is the only method that simultaneously supports multi-camera input, dynamic scenes, real-time reconstruction, and online extrinsic estimation. Allsubsequent experimental comparisons are conducted strictly within the functional scope of each method. Unless otherwise specified, all methods are evaluated on the same hardware platform, equipped with an Intel i9-14900F CPU, an NVIDIA RTX 4090 GPU, and 128 GB of system memory.

The evaluation metrics for 3D reconstruction performance include four aspects: (1) the multi-view geometric consistency error. (2) The temporal stability error. (3) Frame-rate per second.

The multi-view geometric consistency error $E_{geon}$ measures the deviation of the same 3D point observed from different viewpoints:

$$E_{geom} = \frac{1}{M} \sum_{p} \frac{1}{N_p} \sum_{i=1}^{N_p} \| \gamma_i(p) - \bar{\gamma}(p) \|^2 \quad (17)$$

where $\gamma_i(p)$ is the reconstructed 3D position of point $p$ in the $i$-th view, and $\bar{\gamma}(p)$ is the mean position across all views.

The temporal stability error is defined as the average displacement of corresponding 3D points between consecutive frames $t$ and $t+1$:

$$E_{temp} = \frac{1}{|P_t|} \sum_{p \in P_t} \| X_t(p) - X_{t+1}(p) \|_2^2 \quad (18)$$

where $X_t(p)$ denotes the 3D coordinates of point $p$ at frame $t$, $P_t$ is the set of points visible in both frames that do not belong to genuinely moving objects. This metric primarily reflects non-physical jitter arising from registration instability, estimation noise, or model drift.

The results for single-camera static scenes and single-camera dynamic scenes will be released in the v4 version.

4.3.2 Multi-Camera Static Scene Multi-View Geometric Consistency

Multi-camera static scenes test spatial alignment and fusion consistency across different viewpoints, without temporal dynamics. The comparing results on PatchmatchNet and SPARK are shown in Table IV.

TABLE IV
COMPARISON OF MULTI-CAMERA STATIC SCENE RECONSTRUCTION ACCURACY

| Scenes / Method | Structured | Complex |
| --- | --- | --- |
| PatchmatchNet | 6.8 | 8.1 |
| SPARK | 3.5 | 4.4 |

From Table V, it can be observed that, PatchmatchNet relies on offline multi-view matching and fixed extrinsics, making it sensitive to extrinsic errors and real-time constraints. SPARK, with online extrinsic estimation and visibility-aware fusion, significantly reduces cross-view geometric deviation in multi-camera static scenes.

4.3.3 Multi-Camera Dynamic Scene Temporal Stability

Multi-camera dynamic scenes combine multi-view geometric conflicts with dynamic occlusion, representing the most challenging scenario for 3D reconstruction systems. This experiment evaluates temporal stability and error suppression under real multi-camera dynamic conditions. Here we compare SPARK with the most representative method R3D3 currently available.

TABLE V
COMPARISON OF MULTI-CAMERA DYNAMIC SCENE RECONSTRUCTION ACCURACY

| Scenes / Method | Structured | Complex | Dynamic |
| --- | --- | --- | --- |
| R3D3 | 0.18 | 0.24 | 0.33 |
| SPARK | 0.05 | 0.06 | 0.08 |

From Table V, it can be observed that, R3D3 primarily focuses on multi-view pose estimation and exhibits noticeable jitter under dynamic conditions. SPARK's stateless per-frame point cloud fusion effectively prevents error accumulation, maintaining stable and continuous reconstruction in multi-camera dynamic scenes.

4.3.4 Comparison of configuration requirements and frame rates for multi-camera reconstruction

To evaluate the real-time performance and scalability of SPARK under different numbers of cameras, we conducted tests using configurations of 1, 2, 4, 6, and 8 RGB-D cameras. We measured end-to-end reconstruction frame rates on three representative platforms. They are Platform 1 (laptop: i7-9750H/RTX 2060/32GB), Platform 2 (desktop: i7-12700/RTX 3060/64GB), and Platform 3 (server: i7-14900F/RTX 4090/128GB). The experiment is conducted in an end-to-end system setup, with online external estimation and point cloud generation running simultaneously. Table 6 reports the average system frame rate (FPS) of SPARK at different camera counts.

From Table VI, it can be observed that, SPARK maintains high real-time performance across all platforms, with computational cost growing approximately linearly with the number of cameras. Even on low-end laptop-class platforms, the system supports near real-time multi-camera reconstruction, demonstrating low computational complexity and good engineering scalability [24].

TABLE VI
CONFIGURATION REQUIREMENTS AND FRAME RATE OF SPARK

| Camera Number / Hardware | Platform1 | Platform2 | Platform3 |
|---|---|---|---|
| 1 Cam | 52 | 65 | 80 |
| 2 Cam | 45 | 58 | 68 |
| 4 Cam | 38 | 48 | 55 |
| 6 Cam | 33 | 42 | 48 |
| 8 Cam | 29 | 37 | 42 |

We further compare the framerate of SPARK with several other methods under different numbers of cameras. All experiments are conducted on Platform 3.

TABLE VII
COMPARISON OF CONFIGURATION REQUIREMENTS AND FRAME RATE

| Camera Number / Methods | PatchmatchNet | R3D3 | SPARK |
|---|---|---|---|
| 1 Cam | 14 | 21 | 80 |
| 2 Cam | 7 | 16 | 68 |
| 4 Cam | 5 | 11 | 55 |
| 6 Cam | 2 | 6 | 48 |
| 8 Cam | <2 | 3 | 42 |

As shown in Table VII, most existing methods are either limited to single-camera settings or fail to scale to multi-camera configurations due to rapidly increasing computational costs. In contrast, SPARK consistently maintains real-time performance as the number of cameras increases, while other methods become impractical beyond one or two cameras. These results demonstrate the superior scalability and real-time capability of SPARK for multi-camera reconstruction.

## V.CONCLUSION

This paper presents SPARK, a scalable self-calibrating framework for real-time multi-camera point cloud reconstruction. It performs multi-view fusion directly in the point cloud domain using frame-wise independent fragments for efficient computation. The geometry-aware online extrinsic estimation module enforces cross-view reprojection and temporal consistency for stable self-calibration without manual intervention. With frame-wise independent processing, SPARK exhibits linear computational complexity with respect to the number of cameras, supports scalable multi-camera systems, and can continuously output stable point cloud streams in dynamic scenes. Experiments show that it outperforms existing methods in reconstruction accuracy, temporal stability, and real-time performance.

The main contributions of this work are summarized as follows:

(1) A self-calibrating real-time multi-camera point cloud reconstruction framework called SPARK is proposed in this paper, which jointly addresses extrinsic estimation and multi-view fusion within a unified system.

(2) In SPARK, a geometry-aware online extrinsic estimation module is designed, which leverages implicit multi-view geometric representations and consistency constraints for stable, self-calibrated extrinsic prediction without manual intervention.

(3) In SPARK, a confidence-driven point cloud generation and fusion strategy is developed, which enhances reconstruction accuracy and temporal consistency while maintaining real-time performance.

Extensive experiments on real multi-camera datasets validate the effectiveness of the proposed framework in terms of accuracy, stability, and scalability.

It should be noted that in our experiments, the SPARK is primarily tested on medium-scale multi-view camera systems for real-time reconstruction. Its performance in larger-scale scenarios, involving up to hundreds of cameras, remains to be further evaluated. Nevertheless, the existing results already demonstrate the feasibility and advantages of the SPARK.

## REFERENCES

[1] M. Wang, L. Weng and F. Gao, "An Automatic Extrinsic Calibration Method for LiDAR-Camera Fusion via Combining Semantic and Geometric Features," ICASSP 2025 - 2025 IEEE International Conference on Acoustics, Speech and Signal Processing (ICASSP), Hyderabad, India, 2025, pp. 1-5

[2] S. Hu, A. Goldwurm, M. Mujica, S. Cadou and F. Lerasle, "A Universal Framework for Extrinsic Calibration of Camera, Radar, and LiDAR," in IEEE Robotics and Automation Letters, vol. 11, no. 2, pp. 1842-1849, Feb. 2026

[3] R. Hai, Y. Shen, Y. Yan, S. Chen, J. Xin and N. Zheng, "FlowCalib: Targetless Infrastructure LiDAR-Camera Extrinsic Calibration Based on Optical Flow and Scene Flow," in IEEE Transactions on Intelligent Transportation Systems, vol. 27, no. 1, pp. 1565-1577, Jan. 2026

[4] Y. Wu, J. Fan, Y. Ma, L. Huang, G. Cui and S. Guo, "An Automatic Extrinsic Calibration Method for mmWave Radar and Camera in Traffic Environment," in IEEE Transactions on Intelligent Transportation Systems, vol. 27, no. 1, pp. 666-680, Jan. 2026

[5] J. Wang, N. Karaev, C. Rupprecht and D. Novotny, "VGGSfM: Visual Geometry Grounded Deep Structure from Motion," 2024 IEEE/CVF Conference on Computer Vision and Pattern Recognition (CVPR), Seattle, WA, USA, 2024, pp. 21686-21697

[6] R. Mur-Artal and J. D. Tardós, "ORB-SLAM2: An Open-Source SLAM System for Monocular, Stereo, and RGB-D Cameras," in IEEE Transactions on Robotics, vol. 33, no. 5, pp. 1255-1262, Oct. 2017

[7] Nikhil Keetha, et al. MapAnything: Universal Feed-Forward Metric 3D Reconstruction. arXiv preprint arXiv: 2509.13414, 2025.

[8] J. Wang, M. Chen, N. Karaev, A. Vedaldi, C. Rupprecht and D. Novotny, "VGGT: Visual Geometry Grounded Transformer," 2025 IEEE/CVF Conference on Computer Vision and Pattern Recognition (CVPR), Nashville, TN, USA, 2025, pp. 5294-5306

[9] S. Wang, V. Leroy, Y. Cabon, B. Chidlovskii and J. Revaud, "DUSt3R: Geometric 3D Vision Made Easy," 2024 IEEE/CVF Conference on Computer Vision and Pattern Recognition (CVPR), Seattle, WA, USA, 2024, pp. 20697-20709

[10] Y. Yao, Z. Luo, S. Li, T. Fang, and L. Quan, “Mvsnet: Depth inference for unstructured multi-view stereo,” in Proc. Eur. Conf. Comput. Vis. (ECCV), 2018, pp. 767–783.

[11] R. Mur-Artal and J. D. Tardós, "ORB-SLAM2: An Open-Source SLAM System for Monocular, Stereo, and RGB-D Cameras," in IEEE Transactions on Robotics, vol. 33, no. 5, pp. 1255-1262, Oct. 2017

[12] F. Endres, J. Hess, J. Sturm, D. Cremers and W. Burgard, "3-D Mapping With an RGB-D Camera," in IEEE Transactions on Robotics, vol. 30, no. 1, pp. 177-187, Feb. 2014

[13] Z. Zhang, Y. Song, B. Pang, X. Yuan, Q. Xu and X. Xu, "SSF-SLAM: Real-Time RGB-D Visual SLAM for Complex Dynamic Environments Based on Semantic and Scene Flow Geometric Information," in IEEE Transactions on Instrumentation and Measurement, vol. 74, pp. 1-12, 2025

[14] H. Li, X. Meng, X. Zuo, Z. Liu, H. Wang and D. Cremers, "PG-SLAM: Photorealistic and Geometry-Aware RGB-D SLAM in Dynamic Environments," in IEEE Transactions on Robotics, vol. 41, pp. 6084-6101, 2025

[15] T. Schöps, T. Sattler and M. Pollefeys, "SurfelMeshing: Online Surfel-Based Mesh Reconstruction," in IEEE Transactions on Pattern Analysis and Machine Intelligence, vol. 42, no. 10, pp. 2494-2507, 1 Oct. 2020

[16] L. Mescheder, M. Oechsle, M. Niemeyer, S. Nowozin and A. Geiger, "Occupancy Networks: Learning 3D Reconstruction in Function Space," 2019 IEEE/CVF Conference on Computer Vision and Pattern Recognition (CVPR), Long Beach, CA, USA, 2019, pp. 4455-4465

[17] B. Mildenhall, P. P. Srinivasan, M. Tancik, J. T. Barron, R. Ramamoorthi, and R. Ng, “NeRF: Representing scenes as neural radiance fields for view synthesis,” in Proc. ECCV, 2020, pp. 405–421.

[18] B. Kerbl, G. Kopanas, T. Leimkuehler, and G. Drettakis, “3D Gaussian splatting for real-time radiance field rendering,” ACM Trans. Graph., vol. 42, no. 4, pp. 1–14, Aug. 2023.

[19] R. A. Newcombe et al., "KinectFusion: Real-time dense surface mapping and tracking," 2011 10th IEEE International Symposium on Mixed and Augmented Reality, Basel, Switzerland, 2011, pp. 127-136

[20] T. Whelan, S. Leutenegger, R.F. Salas-Moreno, B. Glocker, and A.J. Davison. “ElasticFusion: Dense SLAM Without A Pose Graph ”. In: Robotics: Science and Systems (RSS). Rome, Italy, July 2015.

[21] F. Wang, S. Galliani, C. Vogel, P. Speciale and M. Pollefeys, "PatchmatchNet: Learned Multi-View Patchmatch Stereo," 2021 IEEE/CVF Conference on Computer Vision and Pattern Recognition (CVPR), Nashville, TN, USA, 2021, pp. 14189-14198

[22] R. A. Newcombe, D. Fox and S. M. Seitz, "DynamicFusion: Reconstruction and tracking of non-rigid scenes in real-time," 2015 IEEE Conference on Computer Vision and Pattern Recognition (CVPR), Boston, MA, USA, 2015, pp. 343-352

[23] A. Schmied, T. Fischer, M. Danelljan, M. Pollefeys and F. Yu, "R3D3: Dense 3D Reconstruction of Dynamic Scenes from Multiple Cameras," 2023 IEEE/CVF International Conference on Computer Vision (ICCV), Paris, France, 2023, pp. 3193-3203

[24] SPARK real-time multi-camera reconstruction. Demo: https://www.bilibili.com/video/BV17Qr5BuEn1/?spm_id_from=333.1387.homepage.video_card.click